\documentclass{article}
\usepackage[T1]{fontenc} 
\usepackage[utf8]{inputenc} 
\usepackage{ismir,amsmath,cite,url}
\usepackage{graphicx}
\usepackage{color}

\usepackage{lineno}

\title{Automatic Album Sequencing}

\multauthor
{Vincent Herrmann\hspace{0.05cm}$^*$$^1$ \hspace{1cm} Dylan R.\ Ashley\hspace{0.05cm}$^*$$^{1,2}$ \hspace{1cm} J{\"{u}}rgen Schmidhuber\hspace{0.05cm}$^{1,2}$}
{
$^*$ Equal contribution\\
$^1$ The Swiss AI Lab IDSIA/USI/SUPSI, Lugano, Switzerland\\
$^2$ King Abdullah University of Science and Technology, Thuwal, Saudi Arabia\\
{\tt\small vincent.herrmann@usi.ch, dylan.ashley@usi.ch, juergen.schmidhuber@kaust.edu.sa}
}

\def\authorname{V. Herrmann, D. Ashley, and J. Schmidhuber}

\usepackage[bookmarks=false,pdfauthor={\authorname},pdfsubject={\papersubject},hidelinks]{hyperref}

\begin{document}

\maketitle

\begin{abstract}
Album sequencing is a critical part of the album production process.
Recently, a data-driven approach was proposed that sequences general collections of independent media by extracting the narrative essence of the items in the collections.
While this approach implies an album sequencing technique, it is not widely accessible to a less technical audience, requiring advanced knowledge of machine learning techniques to use.
To address this, we introduce a new user-friendly web-based tool that allows a less technical audience to upload music tracks, execute this technique in one click, and subsequently presents the result in a clean visualization to the user.
To both increase the number of templates available to the user and address shortcomings of previous work, we also introduce a new direct transformer-based album sequencing method.
We find that our more direct method outperforms a random baseline but does not reach the same performance as the narrative essence approach.
Both methods are included in our web-based user interface, and this---alongside a full copy of our implementation---is publicly available at \url{https://github.com/dylanashley/automatic-album-sequencing}
\end{abstract}

\section{Introduction}\label{sec:introduction}

Playlist sequencing is the process of taking a collection of music tracks and ordering them so that listening to them in that order produces a desired emotional response in the listener.
When the collection of music tracks together form an album, this process is known as album sequencing.
Despite its importance in producing impactful music albums and playlists~\cite{cunningham2006more,hansen2009mixing}, this sequencing process has received little attention from the artificial intelligence community.

Our previous research~\cite{ashley2024distillation} introduced a way to compress different kinds of media down into an ultra-low dimensional representation.
This representation captures their relevancy to the overarching story induced by the ordering of the collection they belonged to, i.e., their \textit{narrative essence}.
This is accomplished by using neural networks and contrastive learning~\cite{gutmann2010noise,oord2018representation}.
Then, evolutionary algorithms are used to learn a set of template curves and a novel curve-fitting algorithm to fit the narrative essence of new media collections to these template curves.
The above was principally done with music albums from the FMA dataset~\cite{defferrard2016fma}, though it is shown that this applies to other forms of media as well.

There are two key issues with our previous work.
First, our previous method requires knowledge of advanced machine learning techniques, making it inaccessible to many people who perform album sequencing.
Second, it requires a complex pipeline with (1)~a neural network to extract the narrative essence followed by (2)~a separate evolutionary algorithm to learn a set of templates and then (3)~a fitting algorithm to produce a final ordering.
This is a highly complex and particularly problematic setup that does not allow information like the genre of an album to flow between the narrative essence and the final ordering, resulting in genre-agnostic templates.

Here, we address both of the aforementioned issues.
To address the latter issue, we introduce a new approach that replaces the full pipeline with a single Transformer~\cite{schlag2021linear,schmidhuber1992learning,vaswani2017attention}.
While this does not outperform the more complicated pipeline proposed in our previous work, the new simpler pipeline still outperforms a random baseline, making it useful for automatic album sequencing.
Next, to address the former issue, we implement and release a dedicated user-friendly web-based interface that allows a less technically inclined user to run both the narrative essence-based and the new simplified album sequencing approaches on the user's own music.
We release this interface alongside a complete implementation of our approach publicly at \url{https://github.com/dylanashley/automatic-album-sequencing}

In summary, our contributions are as follows: \textbf{(1)}~We introduce a new direct method to perform automatic album sequencing. \textbf{(2)}~We show that, despite the simpler pipelineethod outperforms a random baseline. \textbf{(3)}~We release a web-based user interface tool that makes automatic album sequencing accessible to a less technical audience.

\section{Related Work}\label{sec:related_work}

While there is a considerable body of work looking at music playlist continuation (e.g., \cite{andric2006automatic,maillet2009steerable,chen2012playlist,bonnin2014automated,vall2019order}), work on album or playlist \textit{sequencing} is more sparse.
Even sparser is work that uses deep neural networks to accomplish this.
These gaps persist despite several works showing the sequencing process's importance and complex nature (see \cite{cunningham2006more,hansen2009mixing,aguiar2021platforms}) and an abundance of work on using deep learning for music recommendation applications (e.g., \cite{van2013deep}).
Most of the existing methods that perform playlist sequencing focus either on controlling the similarity of adjacent songs (e.g., \cite{cliff2000hang,sarroff2012modeling,bittner2017automatic,ikeda2017analysis}) or on directly maximizing the user preference (e.g., \cite{furini2022automatic}).
Altogether, this leaves the most closely related works to ours being our previous work in this area, where we extracted a low-dimensional representation of songs and fit these representations to a set of learned template curves~\cite{ashley2024distillation}, and the concurrent work of Neto et al., who looked at patterns in song-sequencing using pre-defined high-level features~\cite{neto2024algorithmic}.

\section{User Interface}\label{sec:user_interface}

\begin{figure}[ht]
    \centering
    \fbox{\includegraphics[width=.97\linewidth]{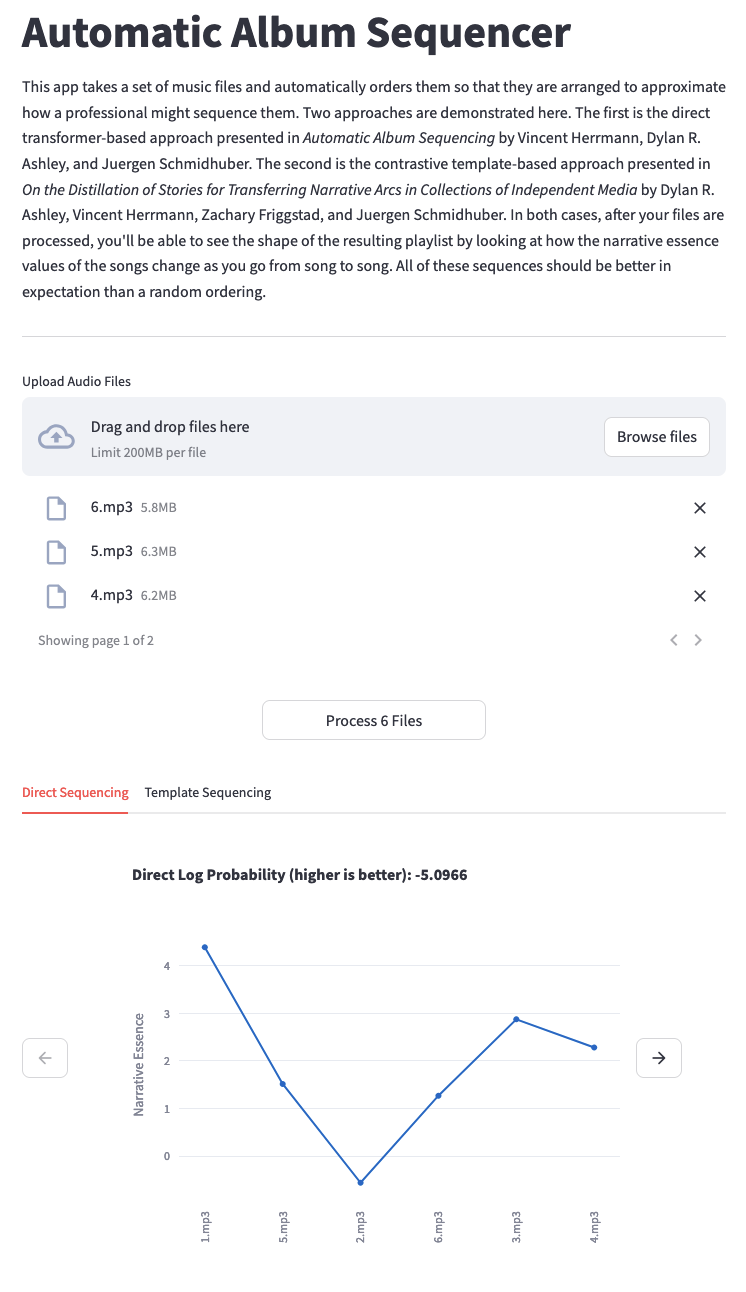}}
    \caption{
        We include a full implementation of our method alongside a clean web-based user interface to make our work more accessible to a less technical audience.
        This implementation is available at \url{https://github.com/dylanashley/automatic-album-sequencing}}
    \label{fig:interface}
\end{figure}

Our first contribution is the web-based user interface shown in Figure~\ref{fig:interface}.
This interface is built in Python\footnote{\url{https://www.python.org/}} using the \texttt{streamlit} library\footnote{\url{https://streamlit.io/}} and thus requires comparatively minimal effort to use.
The interface allows a user to upload files, produce several fittings, and then visualize the results of these fittings.

\section{Direct Album Sequencing}\label{sec:direct_album_sequencing}

\subsection{Method}\label{sec:method}

Let $x_i, i \in 1 \dots M$, represent the $M$ songs of an album $A$, in their original order $o^{(A)}$.
An encoder network $f_\theta$ transforms each song $x_i$ into a low-dimensional representation $z_i$.
The core of our method is the ordering predictor $h_\phi$, whose task is to reconstruct the original album order from the unordered set $\{z_i | 1 \leq i \leq M\}$.

\begin{figure*}
    \centering
    \includegraphics[width=\linewidth]{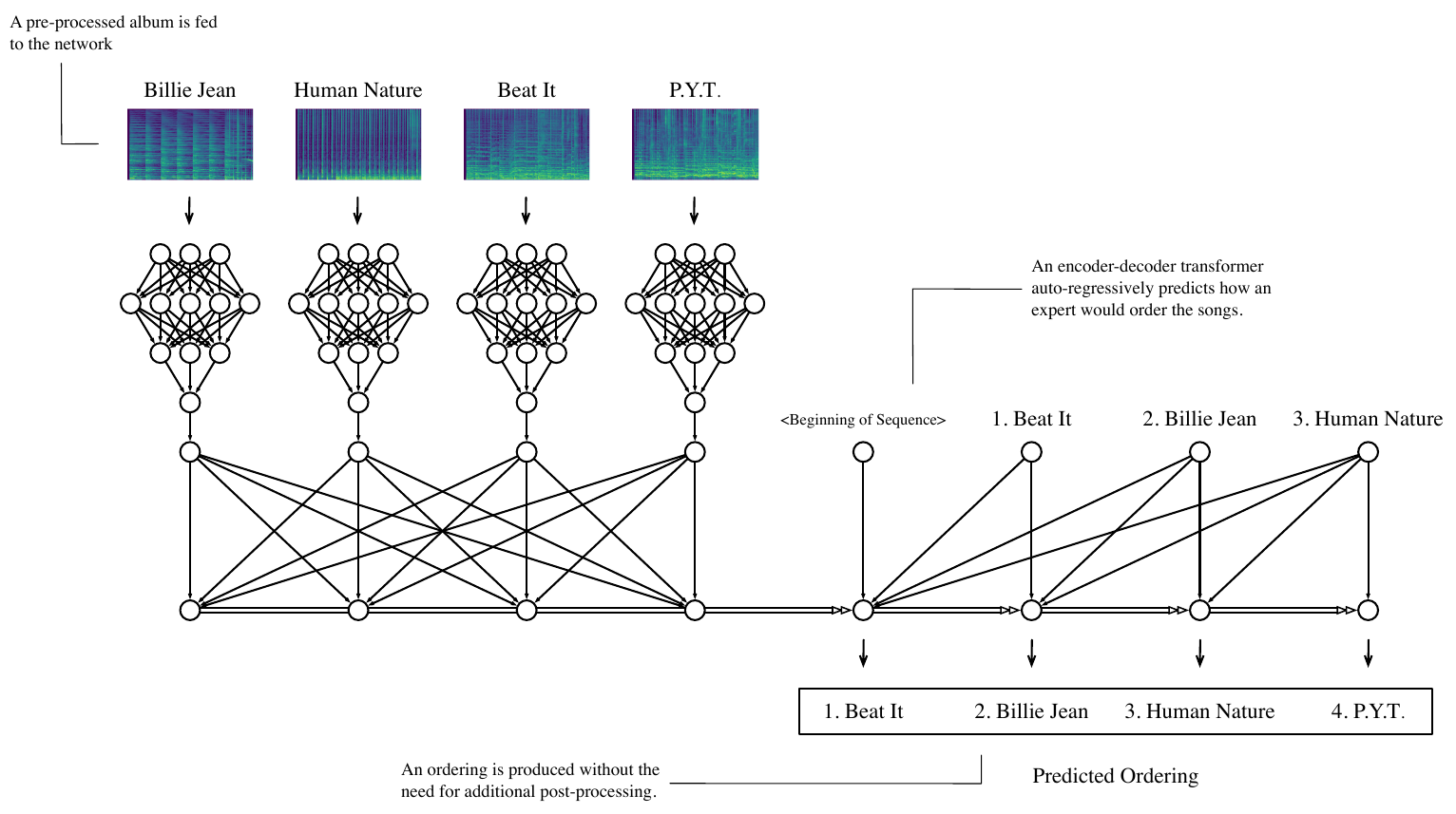}
    \caption{
        Our direct automatic album sequencer is built using a transformer architecture~\cite{schlag2021linear,schmidhuber1992learning,vaswani2017attention}.
        Like in the work of Ashley and Herrmann et al., the input to the network is a shuffled set of preprocessed tracks.}
    \label{fig:ordering_network}
\end{figure*}

We use a sequence-to-sequence model for $h_\phi$ (see Figure~\ref{fig:ordering_network}).
For each album, the input to $h_\phi$ is a sequence of $z_i$ values ordered according to a random permutation $\sigma$.
This permutation $\sigma$ changes with every album and is not known to the model.
The decoder's task is to predict the inverse permutation $\sigma^{-1}$, represented by the indices that reorders the permuted songs into their original album order $o^{(A)}$.

Alternatively, we can frame this task probabilistically.
The model $h_\phi$ estimates the probability $p(o^{(A)} |\{f_\theta(x) | x \in A\})$.
We jointly train the encoder $f_\theta$ and the ordering predictor $h_\phi$ using the standard cross-entropy loss for sequence modelling, where the target is the inverse permutation of the song indices.

In our experiments, $f_\theta$ is a simple two-layer fully connected neural network, and $h_\phi$ is a two-layer encoder-decoder transformer model.
The training data comes from the FMA dataset~\cite{defferrard2016fma}, from which we only take albums of between $3$ and $20$ songs (inclusive).
Each song is represented by a $525$-dimensional feature vector composed of precomputed commonly used attributes provided by the FMA dataset.
The encoder $f_\theta$ reduces every song into a $1$-dimensional feature, similar to the narrative essence of our earlier work~\cite{ashley2024distillation}.
This allows the visualization of the narrative arc (see, e.g., \cite{freytag1894technik,bonds2010spatial,reagan2016emotional}) proposed by the model.
To generate $n$ orders for a single album, we sample $m \geq n$ orders from the distribution modelled by $h_\phi$ and take the top $n$ most likely orders.

\subsection{Results and Discussion}\label{sec:results_and_discussion}

\begin{figure}
    \centering
    \includegraphics[width=\linewidth]{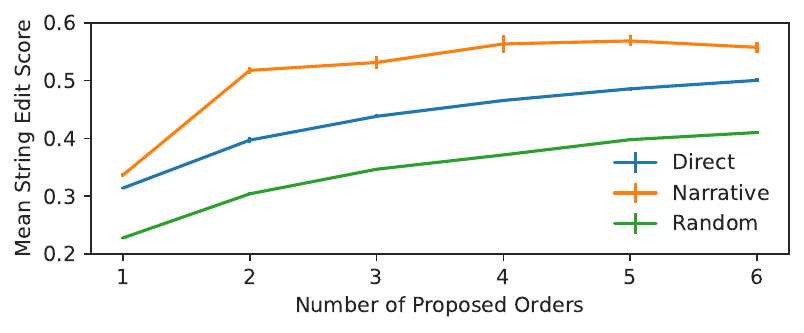}
    \caption{
        While it does not perform as well as the more complicated narrative essence approach, our direct transformer-based approach clearly outperforms a random baseline.
        This means the orders it produces are likewise viable to use for automatic album sequencing.
    }
    \label{fig:results}
\end{figure}

Figure~\ref{fig:results} shows the string edit score of our method as compared to Ashley and Herrmann et al.'s narrative essence approach.
The string edit score measures how closely the set of proposed orders matches the original album order.
It is defined as $f(T, o) = \max_{\hat{o} \in T}{1 - \frac{g(\hat{o}, o)}{|o|}}$, where $T$ is the set of $k$ proposed orders, $o$ is the ground-truth order, and $g$ is the string edit distance (Levenshtein distance~\cite{levenshtein1965binary}).
Despite the much simpler design of ours, for one proposed order, both approaches perform comparably.
For multiple proposed orders, however, our method suffers from not being specifically designed to generate diverse sets of orders.

The direct method is able to capture $1.235 \pm 0.022$ bits of mutual information between album songs and their order, in comparison $1.924 \pm 0.030$ for the narrative essence approach~\cite{ashley2024distillation}.
Also, in the direct method, increasing the dimensionality of the learned representations $z_i$ does not significantly benefit the performance.

\section{Conclusion and Future Work}\label{sec:conclusion_and_future_work}

We introduced a direct transformer-based method for automatic album sequencing and developed a user-friendly web interface to make this technology accessible to a wider audience.
Future work will focus on improving the diversity of generated orderings and incorporating user feedback.

\section{Acknowledgments}\label{sec:acknowledgments}

This work was supported by the Center of Excellence for Generative AI at the King Abdullah University of
Science and Technology (KAUST, Award Number 5940).

\bibliography{main}

\end{document}